\documentclass[conference]{IEEEtran}
\IEEEoverridecommandlockouts
\usepackage{cite}
\usepackage{bm}
\usepackage{amsmath,amssymb,amsfonts}
\usepackage{algorithm}
\usepackage{algorithmic}
\usepackage{graphicx}
\usepackage[letterpaper, left=0.625in, right=0.625in, bottom=1in, top=0.70in]{geometry}

\usepackage{textcomp}
\usepackage{xcolor}
\DeclareMathOperator*{\argmax}{arg\,max}
\newcommand{\be}{\begin{equation}}
\newcommand{\ee}{\end{equation}}

\def\BibTeX{{\rm B\kern-.05em{\sc i\kern-.025em b}\kern-.08em
    T\kern-.1667em\lower.7ex\hbox{E}\kern-.125emX}}

\begin{document}

\title{ Scheduling Policy and Power Allocation for Federated Learning in NOMA Based MEC\\
}
\author{\IEEEauthorblockN{Xiang Ma\IEEEauthorrefmark{1}, Haijian Sun\IEEEauthorrefmark{2}, Rose Qingyang Hu\IEEEauthorrefmark{1}}
\IEEEauthorblockA{
\IEEEauthorrefmark{1}Department of Electrical and Computer Engineering, Utah State University, Logan, UT \\
\IEEEauthorrefmark{2}Department of Computer Science, University of Wisconsin-Whitewater, Whitewater, WI \\
}
}
\maketitle

\begin{abstract}
Federated learning (FL) is a highly pursued  machine learning technique that can train a model centrally while keeping data distributed. Distributed computation makes FL attractive for bandwidth limited applications especially in wireless communications. There can be a large number of distributed edge devices connected to a central parameter server (PS) and iteratively download/upload data from/to the PS. Due to the limited bandwidth, only a subset of connected devices can be scheduled in each round. There are usually millions of parameters in the state-of-art machine learning models such as deep learning, resulting in a high computation  complexity  as well as a high communication burden on collecting/distributing data for training. To improve communication efficiency and make the training model converge faster, we propose a new scheduling policy and power allocation scheme using non-orthogonal multiple access (NOMA) settings to maximize the weighted sum data rate under practical constraints during the entire learning process. NOMA allows multiple users to transmit on the same channel simultaneously. The user scheduling problem is transformed into a maximum-weight independent set problem that can be solved using graph theory. Simulation results show that the proposed scheduling and power allocation scheme can  help achieve a higher FL testing accuracy in NOMA based wireless networks than other existing schemes. 

\end{abstract}


\section{Introduction}
The rapidly growing data availability has gradually enabled training based artificial intelligence applications such as image recognition, autonomous driving, and natural language processing to become reality \cite{big_data}. Unlike the  traditional model based problem solving approaches, machine learning (ML) is more data-driven and less depends on the knowledge of the models. State-of-the-art ML techniques especially deep learning \cite{deep} has demonstrated remarkable performance, such as AlphaGo and Tesla Autopilot, which can outperform human beings in certain areas. Since processing big data may  exceed the computation capability of a single server, processing through multiple distributed \cite{distributed} yet collaborative severs  becomes a highly promising and feasible direction to pursue.   Further motivated by the increasing computational/storage capacities of wireless local devices as well as the ever increasing concerns on sharing data due to privacy and security, next-generation communications/computation networks will encounter a paradigm shift from conventional cloud/central computing to mobile edge computing (MEC) \cite{vehicle}, which largely deploys computational power to the wireless network edge devices to meet the needs of applications that demand very high computations, low latency, as well as high privacy. In this paradigm, a large ML task is partitioned into multiple pieces that can be performed in parallel  by  multiple distributed mobile edge devices based on locally collected data. 

Although data can be processed locally and do not need to be sent in the primitive format to the central parameter server (PS),  data with reduced size may still need to be exchanged for joint processing in order to reach a global consensus on the model learning. Recently, a novel ML technique called federated learning (FL) \cite{federated} is proposed to address this issue. It allows devices to collect data from their local environment and then train models locally. No raw data transmission to the PS is needed. Instead the trained model with a much reduce data size is uploaded to the PS.  There are usually a large number of edge devices connected to one PS.  To achieve efficient learning with limited wireless bandwidth, FL only selects a subset of edge devices for model update in each round. Devices collect data from their respective wireless local environment so the data collected across different devices can be heterogeneous or non-i.i.d.  The significance of user scheduling is to make a decision on selecting a subset of  devices (most important devices based on certain scheduling criteria) to upload model update in each round. The study in \cite{scheduling} gave three different scheduling policies, i.e., random scheduling, round robin, proportional fair to schedule devices randomly, in group and according to channel condition separately. They considered the number of devices and the channel conditions in scheduling but did not consider the data distribution. \cite{coordinated} proposed a coordinated scheduling and power control scheme in cloud radio access networks. To maximize the weight sum data rate, the maximum weight sum data rate problem was transformed to a maximum-weight clique problem. Then the power allocation problem was solved using \cite{mapel} to achieve weighted throughput maximization through power control. It considered user scheduling by using the orthogonal time divsion multiplexing access (TDMA) and frequency division multiplexing access (FDMA). \cite{spectrum} investigated the spectrum efficient resource management problem (SERMP) under non-orthogonal multiple access (NOMA) by transforming the SERMP problem into a maximum weighted independent set problem and solved it using graph theory.

There are usually millions of model parameters in the modern deep learning models such as ResNet, AlexNet.  Most of the  existing works consider a computer-science based methodology to reduce the model size by compression. \cite{robust} utilized quantization and sparsification to perform model compression. Furthermore advanced communication mechanisms have been developed  to improve the spectral efficiency and  to enhance the data rate, which is very instrumental to facilitate the ML methods from communications perspective. When the  transmission takes place in TDMA or FDMA, different devices should work in different time slot or frequency channel.  NOMA  allows multiple devices to transmit simultaneously on the same channel so that data rate is increased and communication latency is reduced when implementing FL \cite{adaptive}. 

In this work, we focus on NOMA based FL uplink communication by considering wireless fading channel. The user scheduling and power allocation are formulated as a maximum weighted sum rate problem, which is further transformed to a maximum weight independent set problem and solved with graph theory. The rest of the paper is organized as follows. Section II introduces the  system model, NOMA transmission scheme and problem formulation. Section III presents the solution for user scheduling and power allocation. Simulation results are shown in Section IV, where experiments are conducted to verify the proposed schemes. Lastly, Section V concludes the paper. 

\section{System Model}
For the distributed learning task on device $k$, there exist a dataset $\bm{x}_k$ and a corresponding label $\bm{y}_k$. At round $t$, ML learns the mapping from $\bm{x}_k^t$ to $\bm{y}_k^t$. Model parameters $\theta_k^t$ are used to describe the mappings. $f(\bm{x}_k, \bm{y}_k; \theta_k^t)$ is the loss function used to capture the error of the mappings. Each user performs the machine learning task locally aiming to solve the following problem \cite{loss_func}:
\be
\min_{\bm{\theta}_k^t}F_k(\bm{\theta}_k^t) = \frac{1}{|\mathcal{D}_k|}  \sum_{i \in \mathcal{D}_k}  f(\mathbf{x}_{k}^t(i), \mathbf{y}_{k}^t(i); \theta_k^t), \label{eq:loss_func_t}
\ee
so we can simply remove the index of $t$ as the equation is true for each  round. 
\be
\min_{\bm{\theta}_k}F_k(\bm{\theta}_k) = \frac{1}{|\mathcal{D}_k|}  \sum_{i \in \mathcal{D}_k}  f(\mathbf{x}_{k}(i), \mathbf{y}_{k}(i); \theta_k), \label{eq:loss_func}
\ee
where $|\mathcal{D}_k|$ is the cardinality of the dataset on user $k$.

FL training relies on the  distributed stochastic gradient descent (DSGD) \cite{sgd} using dataset $\{ \mathcal{D}_1, \mathcal{D}_2, \ldots, \mathcal{D}_K\}$  across $K$ different devices. The  loss function in \eqref{eq:loss_func} can be generalized as: 
\be
\min_{\bm{\theta}}F(\bm{\theta}) =  \sum_{k=1}^K \frac{|\mathcal{D}_k |}{ \mathcal{D}}  F_k (\bm{\theta}_k),
\ee
where $\bm{\theta}$ is the global model that generated from sub-model $\bm{\theta}_k$, $|\mathcal{D}| = \sum_{k=1}^K |\mathcal{D}_k|$. 

\begin{figure}[!h]
    \centering
	\includegraphics[width=65mm, height=29mm]{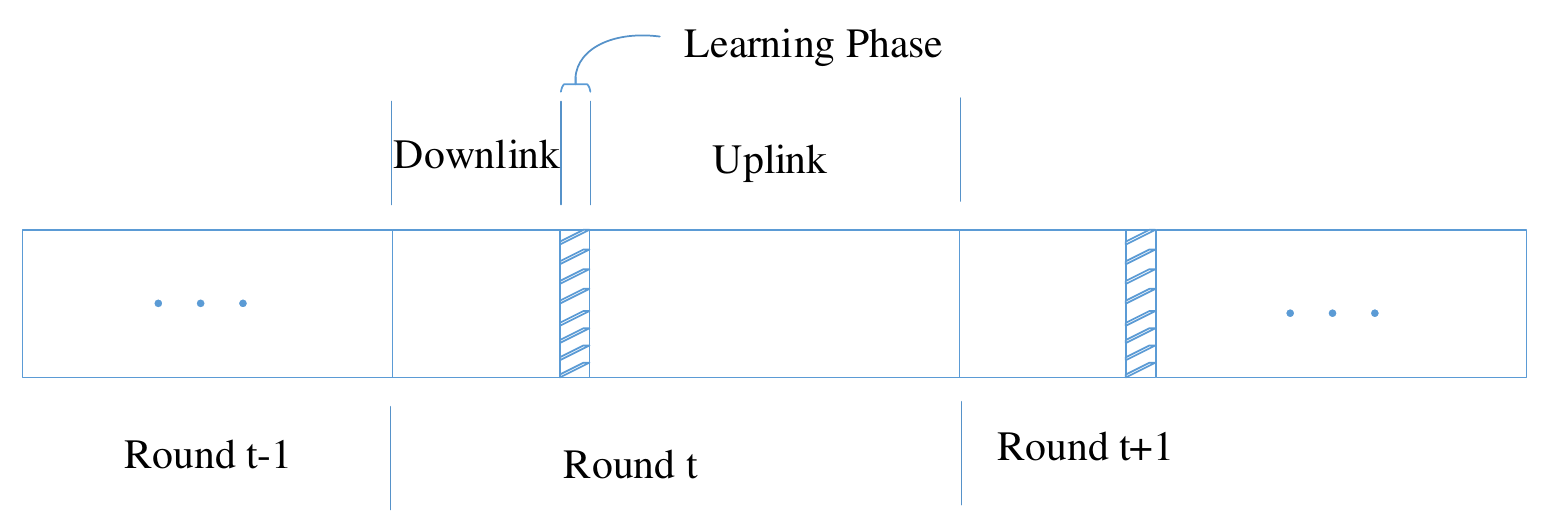}
	\caption{One round of the FL process}
	\label{fig:system_model_phase}
\end{figure}

As shown in Fig. \ref{fig:system_model_phase}, each round of the FL process starts with the downlink communication for sharing central model $\bm{\theta}$, followed by the learning process at local devices to obtain $\bm{\theta}_k$, and ends with the uplink communication from device $k$ to the PS for $\bm{\theta}_k$ transmission.  For uplink, we apply NOMA scheme that allows multiple distributed devices to update simultaneously. 

In our system, there are a total of $M$ edge devices connected to the PS. The maximum number of devices that can be scheduled to participate model update in NOMA is $K$. The total number of iterations or rounds  for the training model to converge is $T$. Let $\mathcal{M}$ be the set of all the devices, $\mathcal{K}$ be the set of devices for model update and $\mathcal{T}$ be the set of all the rounds. Usually the number of devices participating the model updating is much smaller than the total number of devices connected to PS, due to the bandwidth limitation and signaling overhead, i.e., $M \gg K$.  With the existence of massive devices, for the sake of fairness, each device is scheduled to participate the model update at most once. We also assume $M \geq K \times T$.

\begin{figure}[!h]
    \centering
	\includegraphics[width=65mm, keepaspectratio]{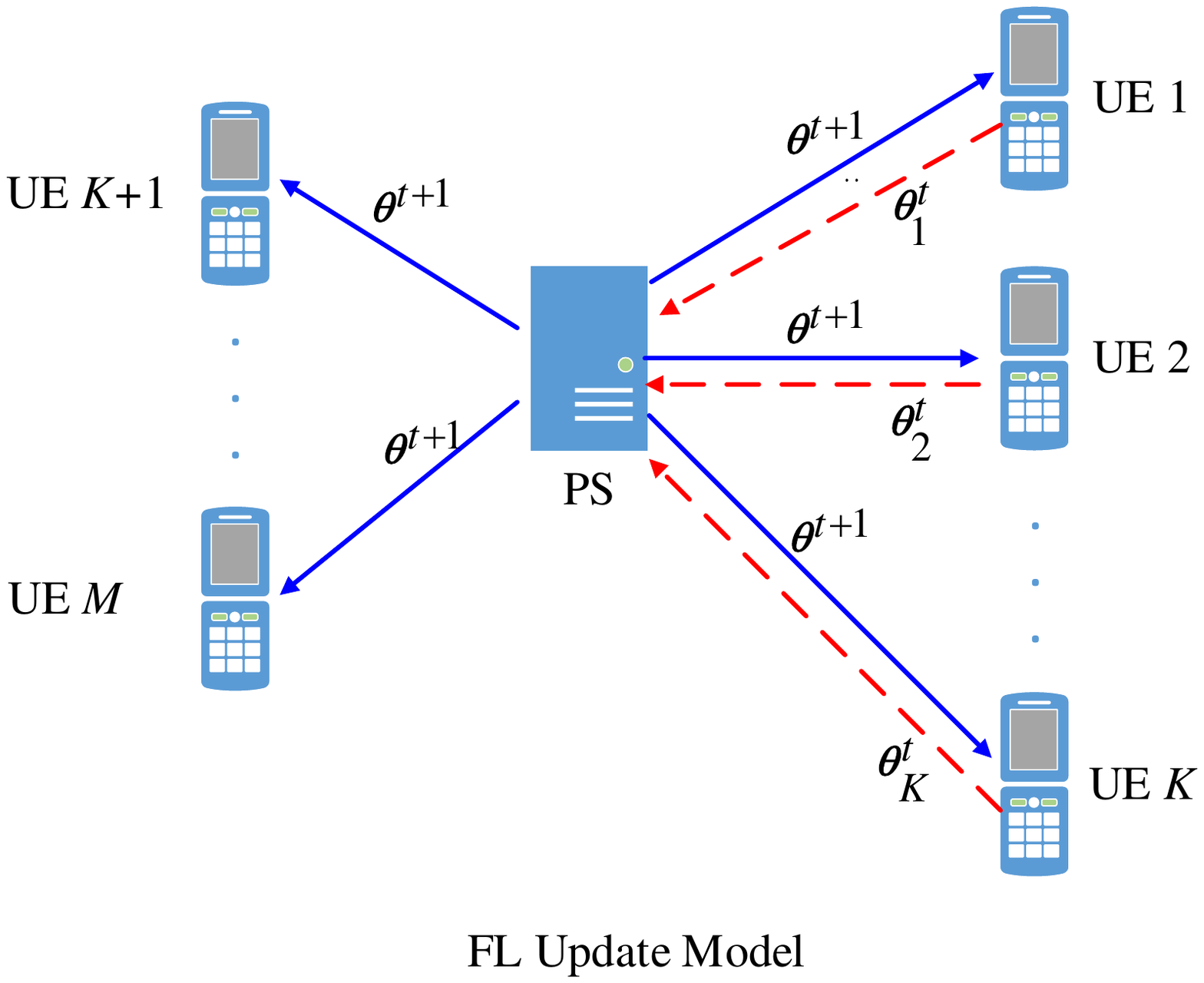}
	\caption{FL Update Model}
	\label{fig:system_model}
\end{figure}

Fig. \ref{fig:system_model} gives the system model of the FL update. At each round, only the right side $K$ devices are scheduled to upload their model update while all the  $M$ devices receive the aggregated model from the PS.
\vspace{-0.15in}

At the beginning, PS initializes the model as $\bm{\theta}^0$ and broadcasts  it to all the users. Each user performs the  local training task and calculates the gradient $\mathbf{g}_k = \nabla F_k(\bm{\theta})$  by using its local data. In the round $t$, user $k$ calculates $\bm{\theta}_k^{t} =  \bm{\theta}_k^{t} - \eta \nabla F_k(\bm{\theta})$ to get gradients $\mathbf{g}_k$, where $\eta$ is the learning rate. All the scheduled users then send their gradients to the PS for aggregation. The PS further calculates $\bm{\theta}^{t+1} = \bm{\theta}^{t} -  \sum_{k=1}^K \mathbf{g}_k$  and  sends $\bm{\theta}^{t+1}$ to all the users for the next round update. This so-called \emph{FedAvg} learning  process continues until the training on the model converges \cite{federated}.

\subsection{Uplink NOMA Transmission}
NOMA allows multiple devices to transmit on the same channel simultaneously. We consider a practical fading channel in typical wireless settings.  The channel gain  of device $k$ at round $t$ is $h_{k}^{t}=L_{k}^{t}h_{0}^t$, which is considered constant during each $t$ but varies across different rounds. $L_{k}^{t}$ is the large-scale fading and $h_0^t$ is the small-scale fading. $L_{k}^{t}$ follows the free-space path loss model $L_k^t = \frac{\sqrt{\delta_k^t} \lambda }{4 \pi d_k^{\alpha/2}}$, $\delta_k^t$ is the transmitter and receiver antenna gain at $t$, $\lambda$ is the signal wavelength, $d_k$ is the distance between user $k$ and the PS, and $\alpha$ is the path-loss exponent. 
Small-scale factor $h_0^t$ is a normal Gaussian variable, i.e.,  $h_0^t \sim \mathcal{CN} (0, 1)$. The transmit power of device $k$ at round $t$ is denoted as $p_{k}^{t}$,  $p_{k}^{t} \leq p_{k}^{tmax}$, where $p_{k}^{tmax}$ is the maximum transmission power. 
Let $s_{k}^t$ be the encapsulated gradient update from user $k$ at round $t$. For simplicity, we normalize the transmitted symbols $ || s_{k}^t  ||_2^2 = 1$. Due to the superposition nature of the transmitted signal in NOMA, the received signal at the PS at round $t$ thus can be expressed as: 
\be
y^t = \sum_{k=1}^K \sqrt{p_k^t} h_{k}^{t} s_k^t + n^t, 
\ee
where $n^t \sim \mathcal{CN} (0, \sigma^2)$ is the additive noise.

NOMA applies successive interference cancellation (SIC) at PS side to decode the signals from different devices sequentially. The decoding process starts with the strongest signal first by regarding other signals as interference. After successful decoding, PS subtracts the decoded signal from the superposed signal and proceeds to decode the next strongest signal. This process continues until the PS decodes all the signals.  Without loss of generality, we assume $p_1^t (h_1^t)^2 > p_2^t (h_2^{t})^2 > \ldots > p_K^t (h_K^t)^2$. Therefore, 
the signal-to-noise-plus-interference ratio (SINR) of user $k$ at round $t$, $\gamma_k^t$ is :
\be
\gamma_k^t = \frac{p_k^t (h_k^t)^2 }{ \sum_{j = k+1}^K p_j^t (h_j^t)^2 + \sigma^2},\forall k = \{1,\ldots,K-1\}.
\ee
The achievable data rate for user $k$ in round $t$ becomes \cite{noma_rate}:
\be
R_k^t = \log_2 \{1+\gamma_k^t \}, \forall k = \{1,\ldots, K-1\}.
\ee
Data rate of the last decoded user $K$ is $R_K^t = \log_2 (1 + \frac{p_K^t (h_K^t)^2}{\sigma^2} )$.

\subsection{Adaptive Model Compression}
Interference exists within each uplink NOMA group, which inevitably impacts the signal quality of different devices. Data rate of each user in a NOMA-based dense wireless network can thus be limited, which may hamper the model update accuracy at each round. A common approach allows each  device to further compress their model to alleviate this limitation. Standard machine learning techniques typically use a $32$-bit floating point number to represent each model parameter. However, the gradients in machine learning tasks are usually in the range  [$-1, 1$] or in  a even smaller range. So less bits can be used to represent the gradients and help reduce the model size. Here limited-bit quantization is applied. DoReFa scheme \cite{dorefa} is suitable for quantizing gradients within [$-1, 1$]. The mapping between full-bit number and less-bit number is established as
\be
q_k(\pi)=\frac{1}{a}\lfloor{a\pi}\rceil.
\ee

$\lfloor\cdot\rceil$ maps to the nearest integer, $\pi$ is the full-bit gradient value, and $a=2^b-1$, where $b$ is the quantization bit length.

Since the data rate of the scheduled devices may vary, we employ adaptive compression to meet different rate limitations. The compression rate $r_k$ for user $k$ can be calculated as $r_k=\max\{\frac{I}{c_k}, 1\}$, $I$ is the total bit length of gradients, $c_k^t = R_k^t t$ is the allowable transmission bit length for user $k$. The quantization bit length $b_k$ is calculated by $b_k=\lfloor\frac{1}{r_k} 32\rfloor$, $\lfloor\cdot\rfloor$ is the floor operation. Further, the compression rate $r_k$ may vary in different rounds, so we can use the average compression rate to represent the compression performance. \textbf{Algorithm 1} summarizes the  proposed compression scheme.
\begin{algorithm}[]
\caption{FL Adaptive Model Compression under NOMA}
\begin{algorithmic}[1]
\STATE  {\bf Initialization:} $\bm{\theta}^0$, $T$. 
\FOR {each FL update round $t$} 
\STATE PS sends $\bm{\theta}^t$ to all users then selects $K$ users. 
\FOR {each selected user $k$ in parallel} 
\STATE {Calculate local gradients:   $\bm{\theta}_k^{t} =  \bm{\theta}_k^{t} - \eta \nabla F_k(\bm{\theta})$. }
\STATE Apply quantization on gradients. 
\STATE Send gradients to the PS. 
\ENDFOR
\STATE PS applies SIC to decode gradient from $K$ users. 
\STATE PS performs weighted average:  $\bm{\theta}^{t+1} = \bm{\theta}^{t} -  \sum_{k=1}^K \frac{|\mathcal{D}_k|}{ \mathcal{D}} \bm{\theta}_k^t$. 
\ENDFOR
\end{algorithmic}
\end{algorithm}
\subsection{Problem Formulation}
Here we provide the formulated optimization problem with  the following three constraints considered in our system model.  

\begin{itemize}
\item $C1$: Each device can be scheduled at most once across different rounds.

\item $C2$: At most $K$ devices are allowed to participate the FL update in each round under NOMA. 

\item $C3$: Transmission power of each device in each round is bounded by a maximum value.

\end{itemize}

We aim to maximize a weighted sum rate of all participated devices, the optimization problem is formulated as
\begin{subequations}
\begin{eqnarray}
& & \max \sum_{m,t} w_m^t \Lambda_m^t R_m^t \label{eq:objective1}\\
s.t. & &   \sum_{t}\Lambda_m^t \leq 1, \forall m,  \label{eq:objective2} \\
     & &  \sum_{m}\Lambda_m^t \leq K, \forall t, \label{eq:objective3} \\
     & &  0 \leq p_m^t \leq {p_m^{tmax}}, \forall(m, t) \in \mathcal{M} \times \mathcal{T}, \label{eq:objective4} \\
& &  \Lambda_m^t \in \{0, 1\}, \forall (m,t) \in \mathcal{M} \times \mathcal{T}, \label{eq:objective6}
\end{eqnarray}
\end{subequations}
where $w_m^t$ is the data rate weight of device $m$ scheduled at round $t$. In FL, PS performs weighted average to generate the current global model, hence a natural selection for the data rate weight can be $w_m^t = \frac{|\mathcal{D}_m |}{ \mathcal{D}}$, which also clearly outlines the significance of each device's update. $\Lambda_m^t = \{ 0, 1\}$ is a binary variable that equals $1$ if device $m$ is scheduled at $t$ and is $0$ otherwise. 
Here, the constraint in \eqref{eq:objective2} corresponds to constraint $C1$, constraint in \eqref{eq:objective3} corresponds to constraint $C2$ and constraint in \eqref{eq:objective4} corresponds to constraint $C3$. Finding the maximum weight sum data rate under these constraints involves traversing all possible scheduling patterns, which possess very high complexity  when the number of total devices is large and selected devices for scheduling is small, i.e.,  $M \gg K$. Towards that, we propose the following scheduling algorithm to address this complexity issue and power allocation to solve the optimization problem \eqref{eq:objective1}. 

\section{Scheduling Algorithm and Power Allocation}

\begin{figure}[!h]
	\centering
	\includegraphics[width=60mm, height=30mm]{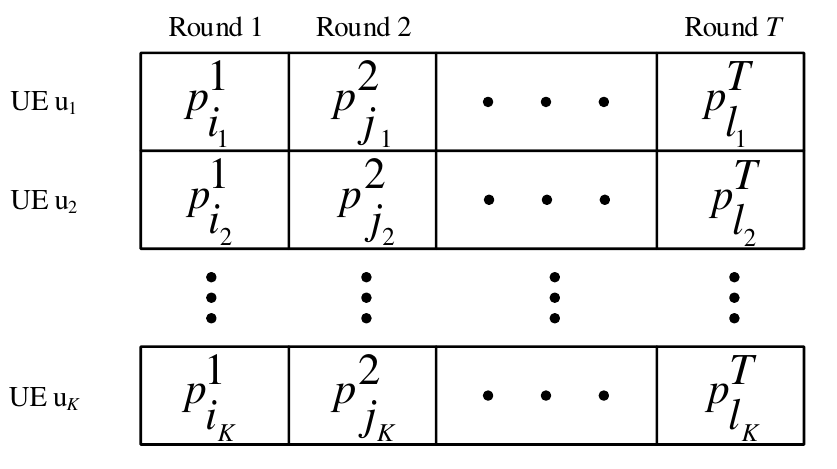}
	\caption{Scheduling Diagram}
	\label{fig:frame_structure}
\end{figure}

Fig. \ref{fig:frame_structure} shows the diagram of the user scheduling. Each column represents a FL round for model update, and there are a total of $T$ columns. Each block in a specific column represents a scheduled user and at most $K$ users are scheduled to participate FL update in each round. The power of the scheduled user $k$ in round $t$ is $p_k^t$. $(i_1, i_2,\dots,i_k)$, $(j_1,j_2,\ldots,j_k)$ and $(l_1,l_2,\dots,l_k)$ are different user combinations.
\vspace{-0.1in}

For the proposed joint scheduling and power allocation scheme, first, all possible user schedules are found. Then optimal power allocation is applied for each schedule to find the optimal one. The scheduling problem which aims to maximize weighted sum rate is transformed under graph theory. Specifically, we introduce the maximum weight independent set problem first. An independent set is a sub-graph of an undirected graph where there exists no edge between any two vertices. When the weight of each vertex is set to be equal to the sum data rate of users scheduled in the specific round, the sum of the weight of all vertices in an independent set equals to the sum data rate of a possible user schedule. The maximum weight independent set then corresponds to the schedule pattern that maximizes the sum data rate. The maximum weight independent set problem involves searching for all possible independent sets and then finding the maximum weight one. Thus a critical step is to construct the scheduling graph in order to find all the scheduling patterns.

\subsection{Scheduling Graph Construction}
\begin{figure}[!h]
    \centering
	\includegraphics[width=55mm]{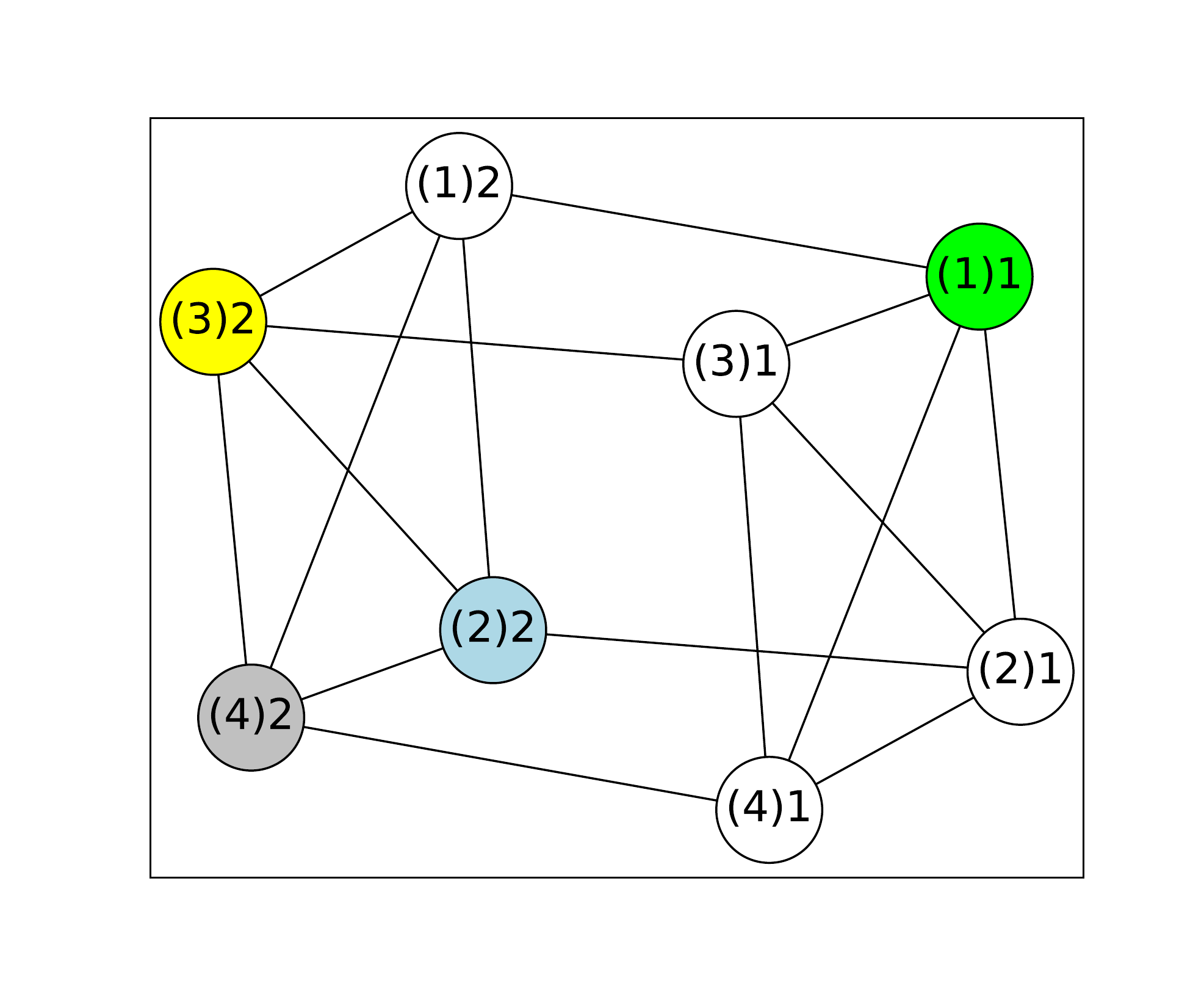}
	\caption{A scheduling graph example}
	\label{fig:scheduling}
\end{figure}

Let $\mathcal{S}$ be the set that includes all the possible scheduling patterns for all  the devices and rounds. $s \in \mathcal{S}$ is a possible schedule. The scheduling graph can be constructed as follows. First, we need to generate vertices. In this graph, a vertex $v_j=(j_1, j_2,\ldots,j_K)t$ indicates that devices $j_1, j_2,\ldots,j_K$ are  scheduled at time $t$. There are a total of ${M \choose K} \times T$ vertices. When creating the edges, the following constraints need to be satisfied. 
\begin{itemize}
\item $C1$: Each device can be scheduled at most once.
\item $C2$: At most $K$ devices can be scheduled in one round.
\end{itemize}
For two vertices $v_i =(i_1, i_2,\dots,i_K)t_i$ and $v_j=(j_1, j_2,\ldots,j_K)t_j$, if $i_k \in \{j_1, j_2,\ldots,j_K\},  \forall k = \{1, \ldots K\}$ (violates $C1$) or $t_i=t_j$ (violates $C2$), $v_i$ and $v_j$ are connected and an edge exists between these two vertices. Then when we select vertices from independent set, both $C1$ and $C2$ will be satisfied. Let us construct a scheduling graph example with $M=4$, $K=1$, and $T=2$,  as shown in Fig. \ref{fig:scheduling}. In this case there are ${4 \choose 1} \times 2=8$ vertices. From this figure, we can find out that the possible independent sets for vertex $(1)1$ (green node) is \{\{$(1)1, (2)2$\}, \{$(1)1, (3)2$\}, \{$(1)1, (4)2$\}\}. Similarly, we can find all the independent sets for each vertex in the graph. Because of the edge connection constraints, each independent set has at most $T$ vertices. Since the FL rounds are continuous and the number of FL rounds is $T$, the independent sets with $T$ vertices are only considered.
\vspace{-0.1in}

\subsection{Optimal scheduling Pattern}
When scheduling graph is constructed, we calculate the weight of each vertex as sum data rate of users scheduled in a specified round, that is 
\be
w(v_j) = \sum_{k \in v_j} w_k^t R_k^t, \forall t \in s. \label{eq:weight}
\ee
Then the sum of the weight of all vertices in an independent set equals the sum data rate of a possible schedule, that is 
\be
\sum_{j}{w(v_j)} = \sum_{k, t} w_k^t R_k^t, \forall(k,t) \in s. \label{eq:sum_weight}
\ee
where $v_j$ represents vertex in an independent set.

The objective function in \eqref{eq:objective1} is actually equal to the problem maximizing the \eqref{eq:sum_weight}, which is the maximum weight independent set problem. The maximum weight sum rate problem then can be transformed as a maximum weight independent set problem. And the optimal schedule can be selected in the \textbf{Algorithm 2}:
\begin{algorithm}[]
\caption{Optimal Scheduling Selection}
\begin{algorithmic}[1]
\STATE  {\bf Require:} $\mathcal{M}, \mathcal{K}, \mathcal{T}, p_m^t$, and $h_m^t$. 
\STATE Initialize {\bf O} = $\O$
\STATE Construct scheduling graph G
\STATE Compute $w(v), \forall v \in G$
\WHILE {$G \ne \O$} \do{
\STATE $Q = \Big \{ v | w(v) \geq \sum_{u \in J(v)} \frac{w(u)}{\beta(u)+1} \Big \}$ 
\STATE Select $v^*=\argmax_{v \in Q} \frac{w(v)}{\beta(v)+1}$
\STATE Set ${\bf O = O} \cup \{v^*\}$
\STATE Set $G=G-J(v^*)$
}
\ENDWHILE
\STATE Output {\bf O}
\end{algorithmic}
\end{algorithm}
here, ${\bf O}$ is the maximum weight independent set in the graph, which is the schedule pattern corresponding maximum weight sum data rate. $J(v)$ is the sub-graph of $G$ containing vertex $v$ and the vertices adjacent to $v$, $\beta(v)$ is the degree of $v$, which is the number of vertices adjacent to $v$. $Q$ is the set of vertices where the weight of vertex $v$ is larger than the average weight of $J(v)$. $v^*$ is selected by making the average weight of $J(v)$ maximization.
\subsection{Power Allocation}
Once the user scheduling is determined,  device power can be allocated according to  the channel condition to achieve the maximum sum data rate. Power allocation in NOMA has  been extensively investigated in the existing works. To achieve the maximum sum data rate under fairness constraints, a similar algorithm to \cite{mapel} is used here. We notice that the objective function \eqref{eq:objective1} as a logarithmic function of SINR is monotonically increasing.  It can be transformed into a product of exponential linear fraction functions. Due to the properties of logarithm function, the optimal power allocation problem for a specified user combination is
\begin{subequations}
\begin{eqnarray}
& &\max \prod \limits_{k=1}^K (\frac{\mu_k(\mathbf{p})}{\phi_k(\mathbf{p})})^{w_k}, \\
s.t. & & 0 \leq p_k\leq p_k^{max}, \forall k \in \mathcal{K}.
\end{eqnarray}
\end{subequations}
where $\mathbf{p}=(p_k, \forall k \in \mathcal{K})$ is the power vector, $\mu_k(\mathbf{p})=\sum_{j=k}^K p_{j}h_{j}^2+\sigma^2$ and $\phi_k(\mathbf{p})=\sum_{j=k+1}^K p_{j} h_{j}^2+\sigma^2$. Let $\mathbf{z}_k=\frac{\mu_k(\mathbf{p})}{\phi_k(\mathbf{p})}$ for all $k$, the problem then can be re-formulated as 
\begin{subequations}
\begin{eqnarray}
& & \max \prod \limits_{k=1}^K (\mathbf{z}_k)^{w_k} \\
s.t. & & 0 \leq \mathbf{z}_k \leq  \frac{\mu_k(\mathbf{p})}{\phi_k(\mathbf{p})} , \forall k \in \mathcal{K}, \\
& & 0 \leq p_k \leq p_k^{max}, \forall k \in \mathcal{K}.
\end{eqnarray}
\end{subequations}
Notice that $\tau(\mathbf{e}) = \prod \limits_{k=1}^K (e_k)^{w_k}$ is an increasing function for all positive $e_k$, where $\mathbf{e}$ is the collection of all $e_k$. Besides, for two vectors $\mathbf{e}_l$ and $\mathbf{e}_m$, if $\mathbf{e}_l \succeq \mathbf{e}_m$, where $\succeq$ means element-wise greater than, we have $\tau(\mathbf{e}_l) > \tau(\mathbf{e}_m)$. Clearly, the optimal solution occurs where $\mathbf{z}_k^* = \frac{\mu_k(\mathbf{p}^*)}{\phi_k(\mathbf{p}^*)}$, and $p_k$ in the feasible set. This can be 
regarded as a multiplicative linear fractional programming (MLFP) problem, where $K$ linear equations are formulated as below: 
\be \label{MLFP}
z_k^*\phi_k(\mathbf{p}^*)-\mu_k(\mathbf{p}^*) = 0, \forall k \in \mathcal{K}.
\ee
Notice that (\ref{MLFP}) contains random channel gain components hence those $K$ linear equations are independent with probability 1, which suggests a unique optimal power allocation  $\mathbf{p}^*$. To solve (\ref{MLFP}) efficiently, however, requires constructing of feasible polyblock and sequentially reduce its size, see \cite{mapel} for the detailed algorithm. 

\section{Simulation results}
This section first gives simulation results to compare two schemes, namely  the TDMA based FedAvg scheme \cite{federated} and NOMA compression based FedAvg scheme. Both schemes use the maximum power transmission for all the devices thus no power control is applied on the uplink. After that we compare the performance for the following four schemes,  1) the scheme using  optimal joint  scheduling with power allocation (our proposed scheme), 2) the scheme using the optimal scheduling but with no power control (all the devices transmit at the maximum power), 3) the scheme  using a random scheduling with optimal power allocation,  and 4) the scheme using  random scheduling with no power control (maximum power transmission). All the simulation runs use image recognition as the learning task trained by the MNIST (Modified National Institute of Standards and Technology) dataset \cite{mnist}. Testing accuracy, which is defined as number of correct predictions divided by total number of predictions, is used to measure the performance of all the schemes. A fully connected neural network called LeNet-$300$-$100$ with two hidden layers is used, which has $300$ neurons in the first layer and $100$ neurons in the second layer. Thus the total number of model parameters is $266,610$. The system parameter settings are given as follows. The uplink bandwidth is $B=4$ MHz, path loss exponent is $\alpha=3$, additive noise power density is $\sigma^2=-174$ dBm/Hz. The total number of user is $M=300$ and the number of model update user in each round is $K=3$. The maximum transmission power of each user is $p^{max}=0.01$ watts. Cell size of PS is $500$ m. Users are uniformly distributed in the cell. Uplink transmission time slot is $t=0.2$ s. For downlink transmission from PS, FL uses broadcast with no compression. Transmission time is $T_d=\max_{k} \frac{I}{B_d \log_2 (1+p_d \gamma_k)}$, where $I$ is the total bit length of model, $B_d$ is the downlink bandwidth and is $10$ MHz. $p_d = 0.2$ watts is the PS transmission power, $\gamma_k$ is the SINR from the PS to $k$-th user.

The hyperparameters are given in Table \ref{Tab:hyper}. The learning phase is partitioned into training and testing stages at each device. Also the dataset are split into training and testing sets correspondingly, which are shown in Table \ref{Tab:hyper}, where  90\% samples belong to the training set and the the rest belong to the testing set. To make the model more general and robust, data are made non-i.i.d across different devices, i.e., the sizes and distributions of data at each device are both different. To evaluate the model validation, in every communication round, each device first does the training based the received model from the PS and local data, followed by the testing process. With iterative  learning, more and more data are fed into the model so that the testing accuracy keeps increasing. 

\begin{table}[h]
	\newcommand{\tabincell}[2]{\begin{tabular}{@{}#1@{}}#2\end{tabular}}
	\centering
	\caption{Hyperparameters\label{Tab:hyper}}
	\begin{tabular}{p{14mm}|p{13mm}|p{13mm}|p{12mm} | p{12mm}}
		\hline
		\textbf{Learning \newline rate size ($\eta$)} & \textbf{Batch \newline size ($\mathcal{B}$)} &
		\textbf{FL \newline Round ($T$)} & \textbf{Training \newline set size} & \textbf{Testing \newline set size}\\
		\hline
		0.01 & 10 & 35 & 90\% & 10\%\\
		\hline
	\end{tabular}
\end{table}

\begin{figure}[!h]
    \centering
	\includegraphics[width=66mm]{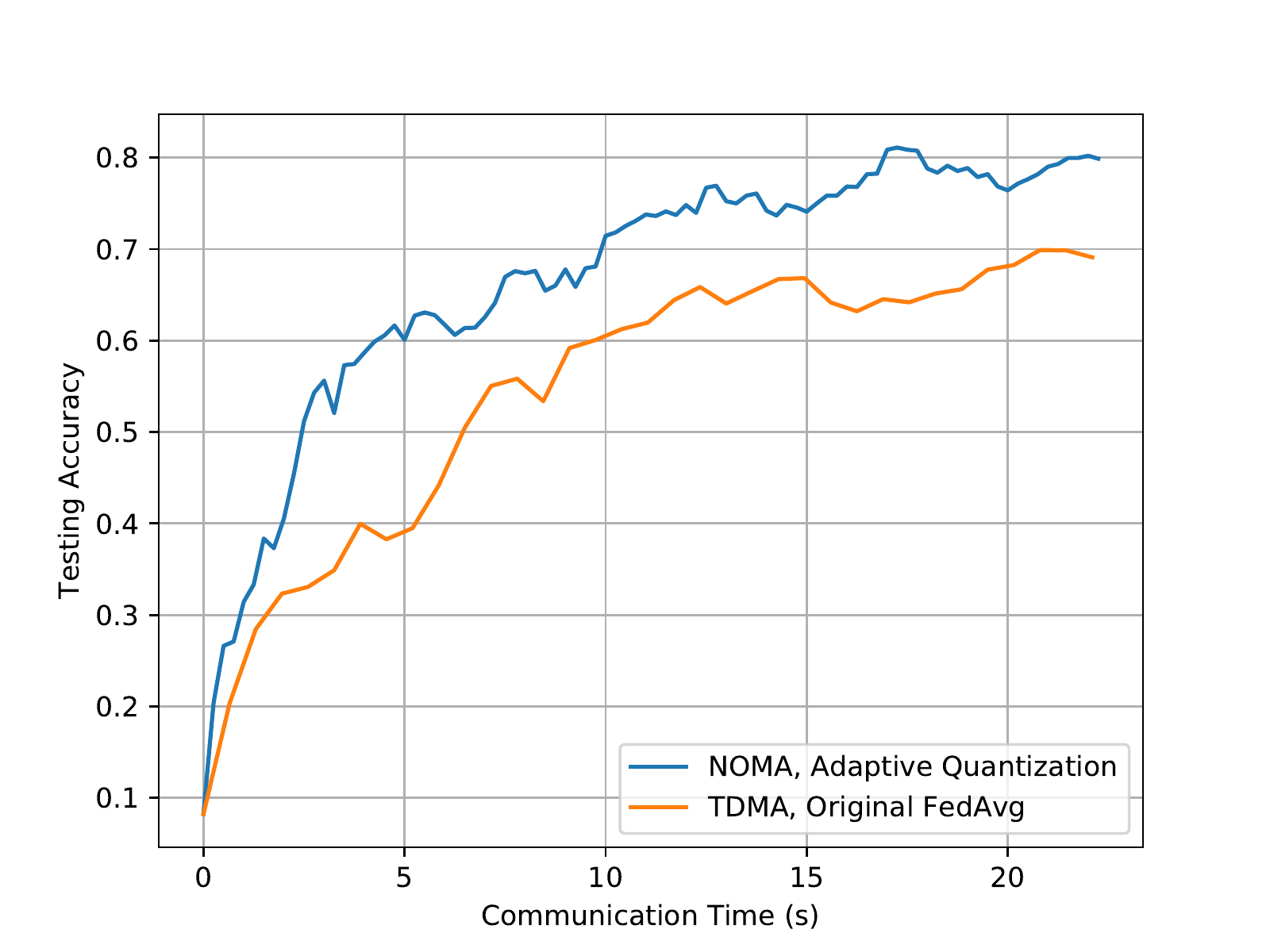}
	\caption{Testing Accuracy vs Communication time}
	\label{fig:tdma_noma}
\end{figure}

We first demonstrate that NOMA compression based FedAvg achieves better performance than the  traditional TDMA based FedAvg. As said, both schemes use the maximum power transmission for all the devices thus no power control is applied on the uplink. In the NOMA based scheme, quantization is used for compression while there is no compression  for the TDMA based scheme. Fig. \ref{fig:tdma_noma} shows that FL using the NOMA based scheme converges faster and achieves a better testing accuracy  compared with the TDMA based scheme. Each  round takes $t_k+T_d$ time in the NOMA based scheme while it takes   $Kt_k+T_d$ time for the TDMA based scheme. So for a given time, NOMA based FedAvg performs more rounds of FL training than the TDMA based FedAvg. In Fig. \ref{fig:tdma_noma}, the NOMA based FedAvg update starts to converge and achieves 70\% of accuracy after $10$s while the TDMA based FedAvg takes about $22$s to achieve the similar accuracy.  
\vspace{-0.20in}
 
\begin{figure}[!h]
	\centering
	\includegraphics[width=66mm]{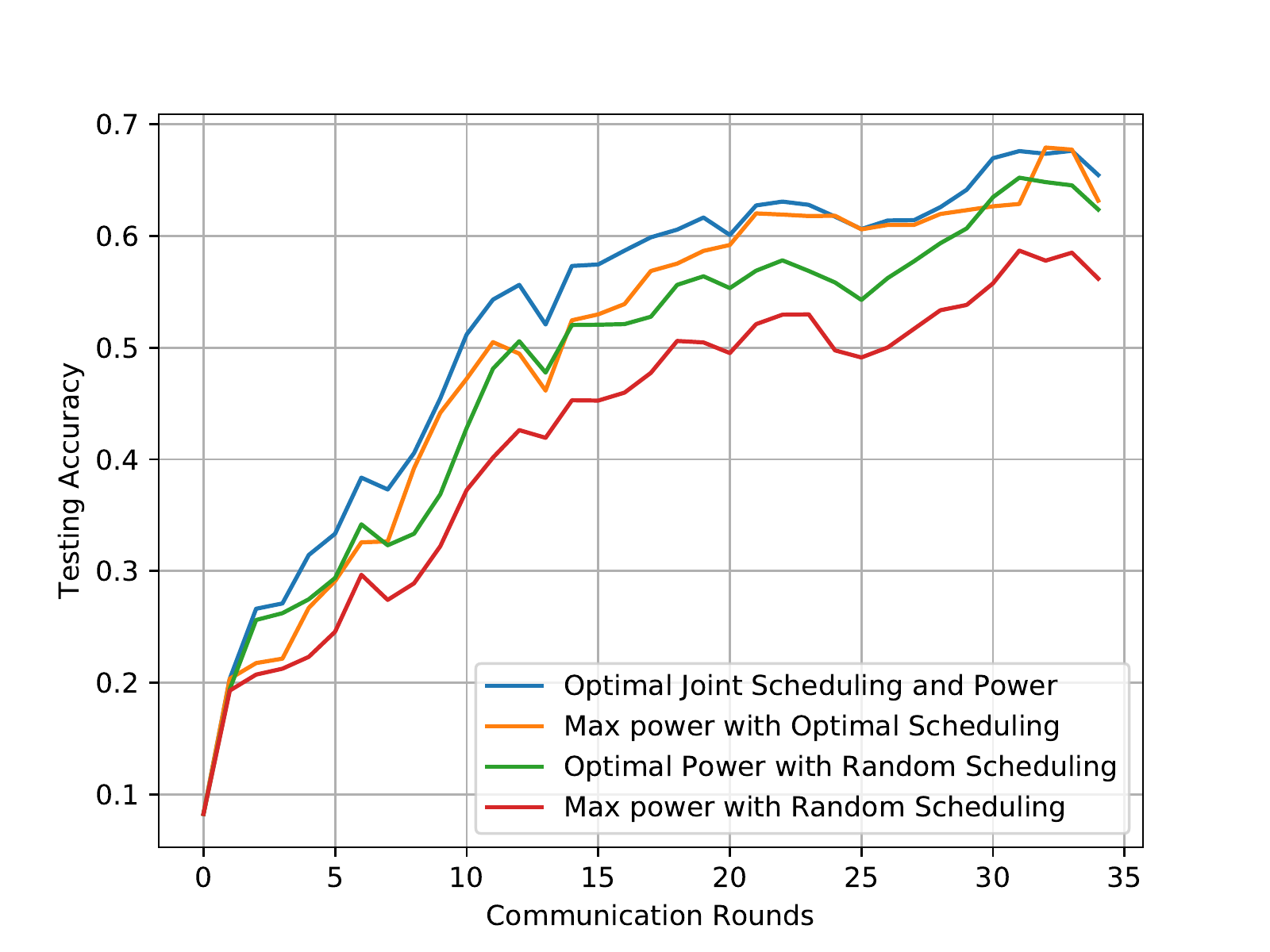}
	\caption{Testing Accuracy vs Communication Rounds}
	\label{fig:max_optimal}
\end{figure}

 Fig. \ref{fig:max_optimal} shows the comparison among 4 different scheduling and power control schemes as defined above. It is observed that all schemes except the 4th one  (random scheduling with maximum power transmission) can get above  $60$\% testing accuracy after $35$  rounds of communication/training.  The  optimal joint scheduling and power allocation scheme consistently achieves the best performance among all the schemes during the entire training process. Both scheduling and power control play an important role in achieving better FL training through improving  the communication quality, which leads to more accurate model update  during the training process. 

\section{Conclusions}
In this work, we proposed to apply NOMA in the FL based model update. To maximize the system sum data rate, the maximum weight sum data rate problem was transformed to a maximum weight independent set problem that can be solved using graph theory based approach. The user scheduling and power allocation were employed to obtain the maximum sum data rate. NOMA based scheme can achieve similar accuracy as TDMA one while reducing the communication latency significantly. Besides, our results show that proper user scheduling and power allocation during wireless communication stage can help to get a higher testing accuracy.

\vspace{12pt}
\end{document}